\documentclass[conference]{IEEEtran}
\IEEEoverridecommandlockouts
\usepackage{cite}
\usepackage{amsmath,amssymb,amsfonts}
\usepackage{graphicx}
\usepackage{textcomp}

\usepackage{booktabs}
\usepackage{amsmath}
\usepackage{graphicx}
\usepackage{algorithm}
\usepackage{algpseudocode}
\usepackage{multirow}
\usepackage{makecell}
\usepackage[table,xcdraw]{xcolor}
\usepackage[most]{tcolorbox}
\usepackage{framed}
\usepackage{float}
\usepackage{url}
\usepackage{hyperref}

\def\BibTeX{{\rm B\kern-.05em{\sc i\kern-.025em b}\kern-.08em
    T\kern-.1667em\lower.7ex\hbox{E}\kern-.125emX}}

\definecolor{brandblue}{rgb}{0.34, 0.7, 1}
\newcounter{promptbox}
\newtcolorbox{promptbox}[1]{%
  enhanced,
  before title=\refstepcounter{promptbox},
  colframe=brandblue,
  fonttitle=\bfseries\footnotesize,
  before upper=\footnotesize,  
  title={Prompt~\thepromptbox:~#1}
}
\newcounter{mainbox}
\newtcolorbox{mainbox}[1]{%
  enhanced,
  before title=\refstepcounter{mainbox},      
  title={Example~\themainbox:~#1},
  colframe=brandblue,
  fonttitle=\bfseries\footnotesize,
  before upper=\footnotesize,
  before skip=0pt,                            
  after  skip=0pt       
}

\begin{document}

\title{Reasoning Capabilities of Large Language Models on Dynamic Tasks\\
}

\author{
  Annie Wong\textsuperscript{*}, 
  Thomas Bäck, 
  Aske Plaat, 
  Niki van Stein, 
  Anna V.~Kononova
  \\
  \vspace{0.2cm} 
  Leiden Institute of Advanced Computer Science, Leiden University, Leiden, The Netherlands
  \\
  \textsuperscript{*}Corresponding author: \texttt{wongasw@vuw.leidenuniv.nl}
}


\maketitle

\begin{abstract}
Large language models excel on static benchmarks, but their ability as self-learning agents in dynamic environments remains unclear. We evaluate three prompting strategies: self-reflection, heuristic mutation, and planning across dynamic tasks with open-source models. We find that larger models generally outperform smaller ones, but that strategic prompting can close this performance gap. Second, an overly long prompt can negatively impact smaller models on basic reactive tasks, while larger models show more robust behaviour. Third, advanced prompting techniques primarily benefit smaller models on complex games, but offer less improvement for already high-performing large language models. Yet, we find that advanced reasoning methods yield highly variable outcomes: while capable of significantly improving performance when reasoning and decision-making align, they also introduce instability and can lead to big performance drops. Compared to human performance, our findings reveal little evidence of true emergent reasoning. Instead, large language model performance exhibits persistent limitations in areas like planning and spatial coordination, suggesting that large language models still suffer fundamental shortcomings that may not be fully overcome through self-reflective prompting alone. Reasoning is a multi-faceted task, and while methods like Chain-of-thought improves multi-step reasoning on math word problems, our findings using dynamic benchmarks highlight important shortcomings in general reasoning capabilities, indicating a need to move beyond static benchmarks to capture the complexity of reasoning. 
\end{abstract}


\begin{figure*}[t]
    \centering    \includegraphics[width=0.78\linewidth,  trim = {0 5cm 0 8cm}]{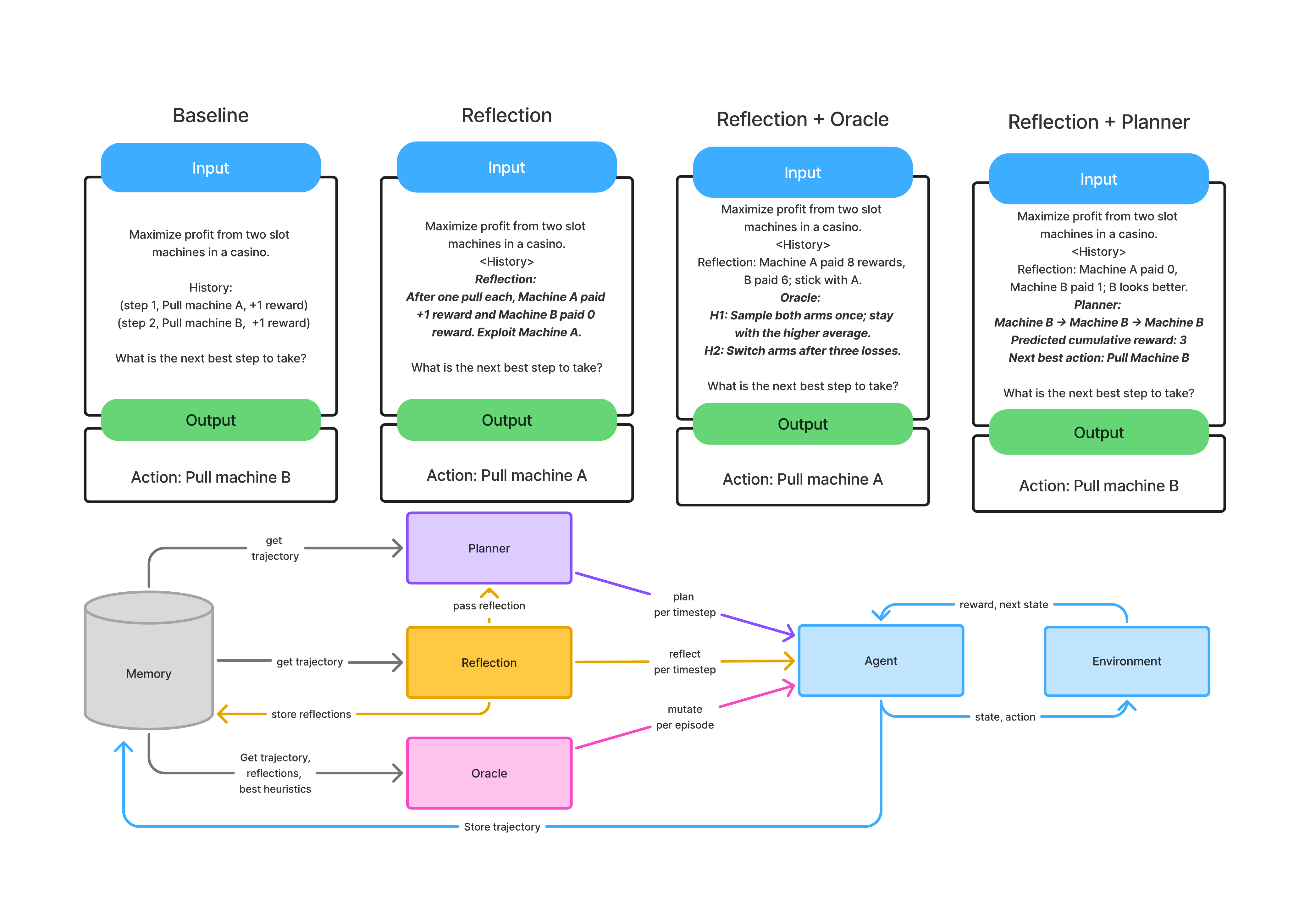}
    \caption{\textbf{Top: }Overview of the four prompting strategies, with each strategy’s added text shown in bold within the prompt. At every timestep in the environment, the agent chooses its next move solely from the text in its prompt: game manual,  history as (state, action, reward, next state) tuples from episode start to the current step, current observation, and the list of legal actions. The example prompts in the figure omit the manual and the action list for brevity, though it is always present at runtime. \textbf{Bottom: }The agent can receive additional guidance from three modules. Reflection performs retrospective analysis based on the past trajectory per timestep. The Oracle evolves heuristics via mutation to uncover environment dynamics based on past reflections and trajectory after each episode. The Planner simulates future states and rewards to recommend the best action based on the past trajectory and current reflection per timestep. }
    \label{fig:agent-framework}
\end{figure*}

\section{Introduction}

A key objective of artificial intelligence has been the development of intelligent agents that can perceive the environment and make autonomous decisions. The emergence of large language models (LLMs) has significantly advanced this objective, showing strong capabilities in the solution of various natural language processing tasks, such as translation, reading comprehension, and answering questions \cite{brown2020language}. These findings show a promising direction towards autonomous agents, and recent studies have begun to explore methods for enabling agents to learn dynamically. For instance, investigations into self-reflection mechanisms allow agents to evaluate their past actions and refine future strategies \cite{shinn2023reflexion}, while work on iterative prompting and environmental feedback aims to facilitate continuous learning \cite{yao2022react}. LLMs have also been investigated to serve as a controller to solve environmental tasks through reasoning and planning \cite{xi2023rise}. The ability of LLM agents to learn in dynamic environments has yet to be definitively proven. While these models excel at in-context learning—generalizing from minimal examples—their reliance on statistical prediction and lack of long-term memory often limit their effectiveness in dynamic settings \cite{tan2023textbased}. Achieving optimal performance on specialised tasks typically requires either fine-tuning with human-annotated data \cite{ouyang2022training} or reliance on careful prompt engineering  \cite{reynolds2021prompt}. These processes are resource-intensive, reducing the flexibility of deploying LLMs in real-world, constantly evolving applications. Specifically, we investigate whether in-context mechanisms can improve continuous learning and multi-step reasoning across various challenges and aim to answer: \textit{To what extent can LLM agents autonomously learn and adapt to tasks in dynamic environments?}". Our contributions are as follows 
\footnote{Code available at https://github.com/ann-w/towards-a-deeper-understanding-of-reasoning-capabilities-in-large-language-models }:


\begin{itemize}
    \item We present a  comparison of open-source LLMs (\textsc{Llama3-8B}, \textsc{Mistral-Nemo-12b}, \textsc{DeepSeek-R1-14b} and \textsc{Llama3.3-70B}) using
    three prompting strategies: \textit{reflection}, \textit{heuristic mutation}, and \textit{planning} across \textsc{SmartPlay} \cite{wu2023smartplay}: a benchmark to evaluate 
    agent capabilities in dynamic tasks. For simple tasks, excessive reasoning harms performance for smaller models, while large models are more robust. Although larger models typically lead in performance, carefully crafted prompts can push smaller models past large-model baselines. These advanced prompting techniques yield the greatest gains for smaller models on complex planning and reasoning tasks, while offering only modest improvements to already high-performing models.
    \item We find that advanced reasoning techniques yield highly variable outcomes: while capable of significantly improving performance when reasoning and decision-making align, they also introduce instability and can lead to big performance drops. 
    \item By transforming a sparse reward into a dense, task-aligned quantitative reward, we demonstrate improvement in the learning effectiveness of LLM agents within complex environments, offering a simpler alternative to the often labour-intensive process of prompt engineering.
    \item We find little evidence for self-learning or emergent reasoning in dynamic tasks that require planning and spatial coordination. These observations highlight the limitations of current, static benchmarks such as question-answer pairs or math word problems, in capturing the complexities of reasoning and revealing inherent deficits.
\end{itemize}

The paper proceeds as follows: Section 2 provides the background, Section 3 details the methodology, Section 4 presents the results, and Section 5 concludes with a discussion of the findings.

\section{Background}
Recent studies have increasingly focused on LLM capabilities as agents in complex, interactive text-based games. These environments present unique challenges requiring puzzle-solving and planning that stretch beyond the inherent statistical prediction capabilities of LLMs. In response, a spectrum of prompt-level techniques has emerged to improve LLM reasoning capabilities without the need for fine-tuning of the parameters.  Chain-of-Thought (CoT) encourages the decomposition of complex tasks into sequential steps \cite{wei2022chain}, while Self-Refine leverages iterative feedback to improve output quality \cite{madaan2024self}. ReAct \cite{yao2022react} and Reflexion \cite{zhang2024agent} integrate reasoning with action and self-reflection, enabling agents to continuously assess and adjust their strategies as they interact with the environment. Others emphasise iterative plan refinement: AutoPlan \cite{ouyang2023autoplan} revises a natural‑language plan while solving household tasks and multi‑hop questions; DEPS \cite{wang2023describe} iteratively refines plans by summarizing situations, diagnosing errors, and selecting sub-goals in open worlds.
Others predict future states and rewards from the interaction history to steer their plan in short‑horizon puzzles \cite{lu2024mental}.
Researchers have also begun to apply evolutionary strategies to self-optimise LLM performance. EvoPrompt \cite{guo2023connecting} uses genetic algorithms and differential evolution to iteratively refine prompts, while metaheuristic methods in frameworks like
LLaMEA \cite{van2024llamea} generate, mutate, and select effective algorithms based on performance feedback. This line of research underscores the potential for evolutionary algorithms to minimize manual tuning. 

Despite the proposal of diverse prompt techniques, methods have usually been studied in isolation. This study offers a comprehensive evaluation by systematically testing different prompting strategies on the SmartPlay suite to test agent capabilities on nine different challenges, making it a more comprehensive benchmark than most studies focusing on a single domain.

\section{Methodology}

\subsection{Agent Framework}
Our agent operates in discrete episodes, each consisting of a sequence of timesteps from a start state to a terminal state. At each timestep $t$, the agent observes the current state and selects an action $a_t$, then receives a reward $r_t$ and transitions to the next state. The objective within a single episode (of horizon $T$ steps) is to maximize the cumulative reward $R = \sum_{k=1}^{T} r_k$, where $T$ is the episode length and $r_k$ is the reward at step $k$. We implement the agent’s policy $\pi$ using a LLM that decides the next action based solely on the information in its prompt at each step. This prompt includes the game’s manual, the task objective, the episode history recorded as a sequence of ⟨state, action, reward, next-state⟩ tuples, the current observation, and the list of possible actions. No parameter fine-tuning is performed; the agent relies on in-context reasoning to choose actions that yield high rewards.

The base agent can be augmented with three modules: Reflection, Oracle, and Planner (Figure 1). Reflection performs a retrospective analysis of the agent’s trajectory at each timestep, providing feedback on recent actions. Oracle operates at the episode level, evolving heuristic rules between episodes to help the agent adapt to changing environment dynamics. Planner is a forward-looking component that simulates future action sequences to recommend the best next action at each timestep. These modules interact with the agent as follows: after the agent takes an action and observes the outcome, the Reflection module (if enabled) generates feedback which is added to the agent’s context; when the Planner is enabled, the agent uses it before deciding an action to explore hypothetical scenarios; and at the end of each episode, if the Oracle is enabled, it updates or mutates a set of heuristics that will be provided to the agent in the next episode.

\subsubsection{Reflection}
\label{sec:reflection}
Adapted from Reflexion \cite{zhang2024agent}, this module enables the agent to analyze its own behavior within the current episode and adjust accordingly. After each action–state transition, the agent inspects the trajectory (stored in memory), compares it to the task objective, and produces a brief suggestion. This reflection can be seen as a meta-level feedback: it compares the agent’s actions against the game’s goals and highlights potential mistakes or better strategies that could lead to a higher reward. We reset the reflection at the start of each new episode, since feedback from a past episode may not be directly applicable when the environment’s conditions change in the next episode. For instance, the profitable bandit might be Bandit 1 in the first episode, but switch to Bandit 2 in the second episode.
\subsubsection{Oracle}
\label{sec:oracle}
The Oracle module aims to capture cross-episode learning by evolving heuristic rules that guide the agent in future episodes using a simple (1+1) evolutionary strategy applied to textual heuristics:  each episode generates one mutated offspring heuristic that replaces the parent only if it achieves a higher reward. After the first episode, the Oracle analyzes the trajectory and reflections from that episode to formulate an initial set of heuristic guidelines. These heuristics are intended to generalize beyond a single episode. For example, a rule might be “Always try the lever that gave a high reward previously” in a multi-armed bandit game. In subsequent episodes, the Oracle continuously refines these rules: at the end of each episode, it generates a mutated version of the current heuristic set (an offspring) by adding, removing, or modifying one of the rules. We then test the offspring in the next episode. If the agent achieves a higher cumulative reward with the new heuristics, the mutated heuristics are kept as the new “parent” set and stored in memory; if not, the Oracle reverts to the previous best heuristics. This evolutionary process allows the agent to adapt to dynamic environments without manual prompt tuning. The Oracle’s heuristics remain fixed during each episode (providing stable guidance within that episode), and only change between episodes as the agent learns from new experience. 
\subsubsection{Planner}
\label{sec:planner}
While Reflection and Oracle focus on the past, the Planner looks ahead to improve decision-making. The Planner simulates possible future trajectories from the current state at each timestep, effectively doing a short-horizon lookahead. Using the game manual, current state, and latest reflection, the Planner simulates rollouts up to three steps, estimates their cumulative rewards, and returns the immediate action with the highest expected return. The Planner’s recommendation for the next action is followed by the agent when the Planner strategy is enabled. While we also experimented with providing Oracle's heuristics to the Planner, we found that the long context caused the LLM to deviate from the prescribed answer format.

\subsection{Models and Hardware Configuration}
We use open source LLMs, including \textsc{Llama3-8B}, \textsc{Mistral-Nemo-12b}, \textsc{DeepSeek-R1-14b} and \textsc{Llama3.3-70B} to investigate how size and architecture affect performance. We ran experiments for the largest model, \textsc{Llama-3.3-70B}, on two A100 machines. Smaller models were run on NVIDIA RTX 4090, RTX 3060, or A40 GPUs. 

\section{Results}
\label{results}


\subsection{Test Environments}

We evaluate our framework on four Smartplay environments\footnote{\url{https://github.com/microsoft/SmartPlay}}.
In Bandit, the agent must balance exploration and exploitation to discover which bandit yields higher payoff. Rock Paper Scissors (RPS) tests probabilistic reasoning against an opponent whose move distribution is biased and randomly shuffled. In Tower of Hanoi, the agent must plan and reason spatially to move three disks across rods, with the rule that a larger disk can never rest on a smaller one. Messenger challenges the agent's ability to understand synonyms and to use that understanding to move around, avoid an enemy, and deliver a message to the goal. We adopt episode counts of \cite{wu2023smartplay}, except that Messenger rollouts are lengthened from 4 to 10 steps because the original horizon often made the goal unreachable, and training is capped at 20 episodes after observing no further performance improvements beyond that point.
\begin{table*}[!tb]
    \centering
    \caption{The table shows the minimum, median and maximum average scores over three runs. The top row shows the human baseline for each game, taken from \cite{wu2023smartplay}. Darker colours indicate closer performance to human baseline. For Bandit, the score improves almost linearly with model size. For simple reactive tasks, excessive reasoning harms small models' performance. Rock Paper Scissors demands adaptation to opponent behaviour in which model size is the main driver of better performance. Reflection + Oracle and Reflection + Planner benefits mostly smaller models in the more complicated Hanoi and Messenger tasks. Yet we observe large variability in runs. $^{\dagger}$ Note that the human baseline is not directly comparable with our results, because it was obtained under a 4-step, 100-episode setting, whereas our evaluation uses a 10-step horizon and 20 episodes.    
    }
    \label{table:scores_table}
    \begin{tabular}{lccccccccccccc}
        \toprule
        Model & Method & \multicolumn{3}{c}{Bandit} & \multicolumn{3}{c}{Rock Paper Scissors} & \multicolumn{3}{c}{Hanoi} & \multicolumn{3}{c}{Messenger} \\
        \midrule
        \textsc{Human Baseline} & - & \multicolumn{3}{c}{\cellcolor[HTML]{C64973}45} & \multicolumn{3}{c}{\cellcolor[HTML]{B73779}43} & \multicolumn{3}{c}{\cellcolor[HTML]{B73779}3} & \multicolumn{3}{c}{\cellcolor[HTML]{B73779}1$^{\dagger}$} \\
        \midrule
         &  & min & med & max & min & med & max & min & med & max & min & med & max \\
        \midrule
        \textsc{llama3-8b} & Base & \cellcolor[HTML]{EC7766}37.45 & \cellcolor[HTML]{DE656B}40.35 & \cellcolor[HTML]{D75D6D}41.65 & \cellcolor[HTML]{FCD29C}16.00 & \cellcolor[HTML]{FCCA96}17.15 & \cellcolor[HTML]{FCC592}17.85 & \cellcolor[HTML]{FCEDB2}0.20 & \cellcolor[HTML]{FCEDB2}0.20 & \cellcolor[HTML]{FCE5AC}0.30 & \cellcolor[HTML]{FCBD8B}-0.45 & \cellcolor[HTML]{FC9A6F}-0.15 & \cellcolor[HTML]{FC9A6F}-0.15 \\
        \textsc{llama3-8b} & Reflection & \cellcolor[HTML]{F27D64}36.40 & \cellcolor[HTML]{ED7766}37.40 & \cellcolor[HTML]{E77068}38.50 & \cellcolor[HTML]{FCD9A2}15.05 & \cellcolor[HTML]{FCD09A}16.40 & \cellcolor[HTML]{FCB887}19.80 & \cellcolor[HTML]{FCDEA5}0.40 & \cellcolor[HTML]{FCB786}0.90 & \cellcolor[HTML]{FCAF80}1.00 & \cellcolor[HTML]{FCB181}-0.35 & \cellcolor[HTML]{F88462}0.05 & \cellcolor[HTML]{F58063}0.10 \\
        \textsc{llama3-8b} & Reflection + Oracle & \cellcolor[HTML]{F47F63}36.00 & \cellcolor[HTML]{EF7965}37.00 & \cellcolor[HTML]{DF676A}40.00 & \cellcolor[HTML]{FCEEB3}12.00 & \cellcolor[HTML]{FC8C63}26.00 & \cellcolor[HTML]{D4596E}36.00 & \cellcolor[HTML]{FCFDBF}0.00 & \cellcolor[HTML]{FCFDBF}0.00 & \cellcolor[HTML]{FCAF80}1.00 & \cellcolor[HTML]{FCFDBF}-1.00 & \cellcolor[HTML]{FC8961}0.00 & \cellcolor[HTML]{FC8961}0.00 \\
        \textsc{llama3-8b} & Reflection + Planner & \cellcolor[HTML]{FCAF80}30.00 & \cellcolor[HTML]{FC8D64}34.00 & \cellcolor[HTML]{F98561}35.00 & \cellcolor[HTML]{FCD9A2}15.00 & \cellcolor[HTML]{FCCB97}17.00 & \cellcolor[HTML]{F58163}28.00 & \cellcolor[HTML]{FCFDBF}0.00 & \cellcolor[HTML]{FCAF80}1.00 & \cellcolor[HTML]{E56D69}2.00 & \cellcolor[HTML]{FC8961}0.00 & \cellcolor[HTML]{FC8961}0.00 & \cellcolor[HTML]{FC8961}0.00 \\
        \midrule
        \textsc{mistral-nemo-12b} & Base & \cellcolor[HTML]{FCA779}30.95 & \cellcolor[HTML]{FC8B63}34.20 & \cellcolor[HTML]{F27E64}36.30 & \cellcolor[HTML]{FCD09A}16.40 & \cellcolor[HTML]{FCCF99}16.50 & \cellcolor[HTML]{FCBD8B}19.05 & \cellcolor[HTML]{FCFDBF}0.00 & \cellcolor[HTML]{FCFDBF}0.00 & \cellcolor[HTML]{FCF5B8}0.10 & \cellcolor[HTML]{FCA073}-0.20 & \cellcolor[HTML]{FCA073}-0.20 & \cellcolor[HTML]{FC9A6F}-0.15 \\
        \textsc{mistral-nemo-12b} & Reflection & \cellcolor[HTML]{FCBC8A}28.50 & \cellcolor[HTML]{FCA275}31.50 & \cellcolor[HTML]{FC9D71}32.15 & \cellcolor[HTML]{FCC28F}18.30 & \cellcolor[HTML]{FCB887}19.70 & \cellcolor[HTML]{FCB383}20.50 & \cellcolor[HTML]{FCE5AC}0.30 & \cellcolor[HTML]{FCD69F}0.50 & \cellcolor[HTML]{FCCE99}0.60 & \cellcolor[HTML]{FCB786}-0.40 & \cellcolor[HTML]{FCB181}-0.35 & \cellcolor[HTML]{FCA678}-0.25 \\
        \textsc{mistral-nemo-12b} & Reflection + Oracle & \cellcolor[HTML]{FCFDBF}21.00 & \cellcolor[HTML]{FCB887}29.00 & \cellcolor[HTML]{E56D69}39.00 & \cellcolor[HTML]{FCD9A2}15.00 & \cellcolor[HTML]{FCC491}18.00 & \cellcolor[HTML]{FC8C63}26.00 & \cellcolor[HTML]{FCFDBF}0.00 & \cellcolor[HTML]{FCAF80}1.00 & \cellcolor[HTML]{FCAF80}1.00 & \cellcolor[HTML]{FCFDBF}-1.00 & \cellcolor[HTML]{FC8961}0.00 & \cellcolor[HTML]{B73779}1.00 \\
        \textsc{mistral-nemo-12b} & Reflection + Planner & \cellcolor[HTML]{FCEBB1}23.00 & \cellcolor[HTML]{FCC995}27.00 & \cellcolor[HTML]{FCC08E}28.00 & \cellcolor[HTML]{FCFDBF}10.00 & \cellcolor[HTML]{FCF5B9}11.00 & \cellcolor[HTML]{E0686A}33.00 & \cellcolor[HTML]{FCFDBF}0.00 & \cellcolor[HTML]{FCAF80}1.00 & \cellcolor[HTML]{FCAF80}1.00 & \cellcolor[HTML]{FCFDBF}-1.00 & \cellcolor[HTML]{B73779}1.00 & \cellcolor[HTML]{B73779}1.00 \\
        \midrule
        \textsc{deepseek-r1-14b} & Base & \cellcolor[HTML]{DD646B}40.55 & \cellcolor[HTML]{DA616C}41.00 & \cellcolor[HTML]{D85F6D}41.40 & \cellcolor[HTML]{FCC894}17.50 & \cellcolor[HTML]{FCBF8C}18.80 & \cellcolor[HTML]{FCBC8A}19.15 & \cellcolor[HTML]{FCCE99}0.60 & \cellcolor[HTML]{FCAF80}1.00 & \cellcolor[HTML]{FCA073}1.20 & \cellcolor[HTML]{EA7467}0.25 & \cellcolor[HTML]{E0686A}0.40 & \cellcolor[HTML]{D65B6E}0.55 \\
        \textsc{deepseek-r1-14b} & Reflection & \cellcolor[HTML]{FC9369}33.30 & \cellcolor[HTML]{FB8861}34.55 & \cellcolor[HTML]{F98561}35.00 & \cellcolor[HTML]{FCBB8A}19.25 & \cellcolor[HTML]{FCB685}20.05 & \cellcolor[HTML]{FCB585}20.15 & \cellcolor[HTML]{FCD69F}0.50 & \cellcolor[HTML]{FCC693}0.70 & \cellcolor[HTML]{FCA073}1.20 & \cellcolor[HTML]{F17C64}0.15 & \cellcolor[HTML]{EA7467}0.25 & \cellcolor[HTML]{E0686A}0.40 \\
        \textsc{deepseek-r1-14b} & Reflection + Oracle & \cellcolor[HTML]{EF7965}37.00 & \cellcolor[HTML]{EA7367}38.00 & \cellcolor[HTML]{DF676A}40.00 & \cellcolor[HTML]{FCD9A2}15.00 & \cellcolor[HTML]{FCB686}20.00 & \cellcolor[HTML]{FCAF80}21.00 & \cellcolor[HTML]{FCAF80}1.00 & \cellcolor[HTML]{FCAF80}1.00 & \cellcolor[HTML]{FCAF80}1.00 & \cellcolor[HTML]{FC8961}0.00 & \cellcolor[HTML]{B73779}1.00 & \cellcolor[HTML]{B73779}1.00 \\
        \textsc{deepseek-r1-14b} & Reflection + Planner & \cellcolor[HTML]{FCB887}29.00 & \cellcolor[HTML]{FC9E72}32.05 & \cellcolor[HTML]{FC956B}33.00 & \cellcolor[HTML]{FCBA88}19.50 & \cellcolor[HTML]{FCB685}20.05 & \cellcolor[HTML]{FCB484}20.25 & \cellcolor[HTML]{FCFDBF}0.00 & \cellcolor[HTML]{FCB786}0.90 & \cellcolor[HTML]{E56D69}2.00 & \cellcolor[HTML]{E0686A}0.40 & \cellcolor[HTML]{B73779}1.00 & \cellcolor[HTML]{B73779}1.00 \\
        \midrule
        \textsc{llama3.3-70b} & Base & \cellcolor[HTML]{DB626C}40.90 & \cellcolor[HTML]{D75D6D}41.70 & \cellcolor[HTML]{D65C6E}41.90 & \cellcolor[HTML]{FCAB7D}21.55 & \cellcolor[HTML]{FCA779}22.20 & \cellcolor[HTML]{FC9369}25.05 & \cellcolor[HTML]{E56D69}2.00 & \cellcolor[HTML]{E56D69}2.00 & \cellcolor[HTML]{E56D69}2.00 & \cellcolor[HTML]{FC946A}-0.10 & \cellcolor[HTML]{F58063}0.10 & \cellcolor[HTML]{F58063}0.10 \\
        \textsc{llama3.3-70b} & Reflection & \cellcolor[HTML]{D85E6D}41.50 & \cellcolor[HTML]{D85E6D}41.50 & \cellcolor[HTML]{D1566F}42.80 & \cellcolor[HTML]{FC8C64}25.95 & \cellcolor[HTML]{FA8761}26.80 & \cellcolor[HTML]{F78362}27.55 & \cellcolor[HTML]{FCC693}0.70 & \cellcolor[HTML]{FCAF80}1.00 & \cellcolor[HTML]{EE7865}1.80 & \cellcolor[HTML]{E36C69}0.35 & \cellcolor[HTML]{E0686A}0.40 & \cellcolor[HTML]{E0686A}0.40 \\
        \textsc{llama3.3-70b} & Reflection + Oracle & \cellcolor[HTML]{E56D69}39.00 & \cellcolor[HTML]{CB4F71}44.00 & \cellcolor[HTML]{CB4F71}44.00 & \cellcolor[HTML]{FCD9A2}15.00 & \cellcolor[HTML]{F17C64}29.00 & \cellcolor[HTML]{ED7766}30.00 & \cellcolor[HTML]{E56D69}2.00 & \cellcolor[HTML]{E56D69}2.00 & \cellcolor[HTML]{E56D69}2.00 & \cellcolor[HTML]{FCFDBF}-1.00 & \cellcolor[HTML]{FC8961}0.00 & \cellcolor[HTML]{B73779}1.00 \\
        \textsc{llama3.3-70b} & Reflection + Planner & \cellcolor[HTML]{FC8D64}34.00 & \cellcolor[HTML]{DF676A}40.00 & \cellcolor[HTML]{B73779}48.00 & \cellcolor[HTML]{F98661}27.00 & \cellcolor[HTML]{ED7766}30.00 & \cellcolor[HTML]{D05470}37.00 & \cellcolor[HTML]{FCAF80}1.00 & \cellcolor[HTML]{FCAF80}1.00 & \cellcolor[HTML]{E56D69}2.00 & \cellcolor[HTML]{FCFDBF}-1.00 & \cellcolor[HTML]{FCFDBF}-1.00 & \cellcolor[HTML]{B73779}1.00 \\
        \bottomrule
    \end{tabular}
\end{table*}

\subsection{Results and Analysis}

Table \ref{table:scores_table}  shows the scores for each strategy. For each combination of strategy and game environment, we ran three independent runs\footnote{All results are averaged over three independent runs; the budget is capped at three because of the computational cost imposed by \textsc{LLaMA3.3-70B}.}. In each run, the agent with a given  strategy played through the fixed number of episodes, and we computed the average score across those episodes for that run. Then, we rank these three average scores. The highest score is reported as the maximum, the lowest score as the minimum, and the middle score as the median. Each environment defines success differently: in Bandit and Rock–Paper–Scissors, the score represents the frequency of optimal actions; in Hanoi, it is the count of moved disks to the goal and in Messenger, it is the accumulated reward, where correct message actions yield +1 and invalid moves incur -1. Human baselines are taken from \cite{wu2023smartplay}. 
In the following sections, we report results as \textit{median score [min, max]}.

\subsubsection{Model Size and Performance}
Larger parameter counts generally achieve higher scores (see Table~\ref{table:scores_table}). For instance, on Bandit, \textsc{Llama3.3-70b} $41.70 \,[40.90-41.90]$ outperforms \textsc{deepseek-r1-14b} $41.00\, [40.55-41.40]$, \textsc{Mistral-Nemo-12b} $34.20\,[30.95-36.30]$ and \textsc{Llama3-8b}  $40.35\,[37.45-41.65]$. Model performance increases with size across all game baselines. This performance gap widens as the games become more challenging. 
Prompting techniques can close this gap on specific tasks. For example, \textsc{Llama3-8b} with Reflection + Oracle $26.00\,[12.00-36.00]$ outperforms \textsc{Llama3.3-70b}’s baseline of $22.20\,[21.55-25.05]$ on RPS. In Messenger, \textsc{Mistral-Nemo-12b} with \textit{Reflection + Planner} $1.00\,[-1.00-1.00]$ exceeds \textsc{Llama3.3-70b}’s baseline $0.10\,[-0.10-0.10]$.
However, these improvements also increase score variance across all models. For instance, \textsc{Mistral-Nemo-12b}'s Reflection + Planner score in RPS ranges from 10.00 to 33.00, and \textsc{DeepSeek-R1-14B}'s Reflection + Planner ranges from 0.00 to 2.00 in Hanoi. This indicates that advanced prompting strategies can yield significant improvements but are also unstable and can lead to big performance drops. Specifically, adding the Planner module on top of Reflection usually widens the variance more than the Reflection + Oracle strategy.  While prompting strategies can offset smaller parameter counts, larger models still offer the most consistent advantage. Achieving performance improvements solely through in-context learning is challenging and often comes with increased variability for all model sizes.

\begin{figure*}[!tb]
  \centering  \includegraphics[width=\textwidth]{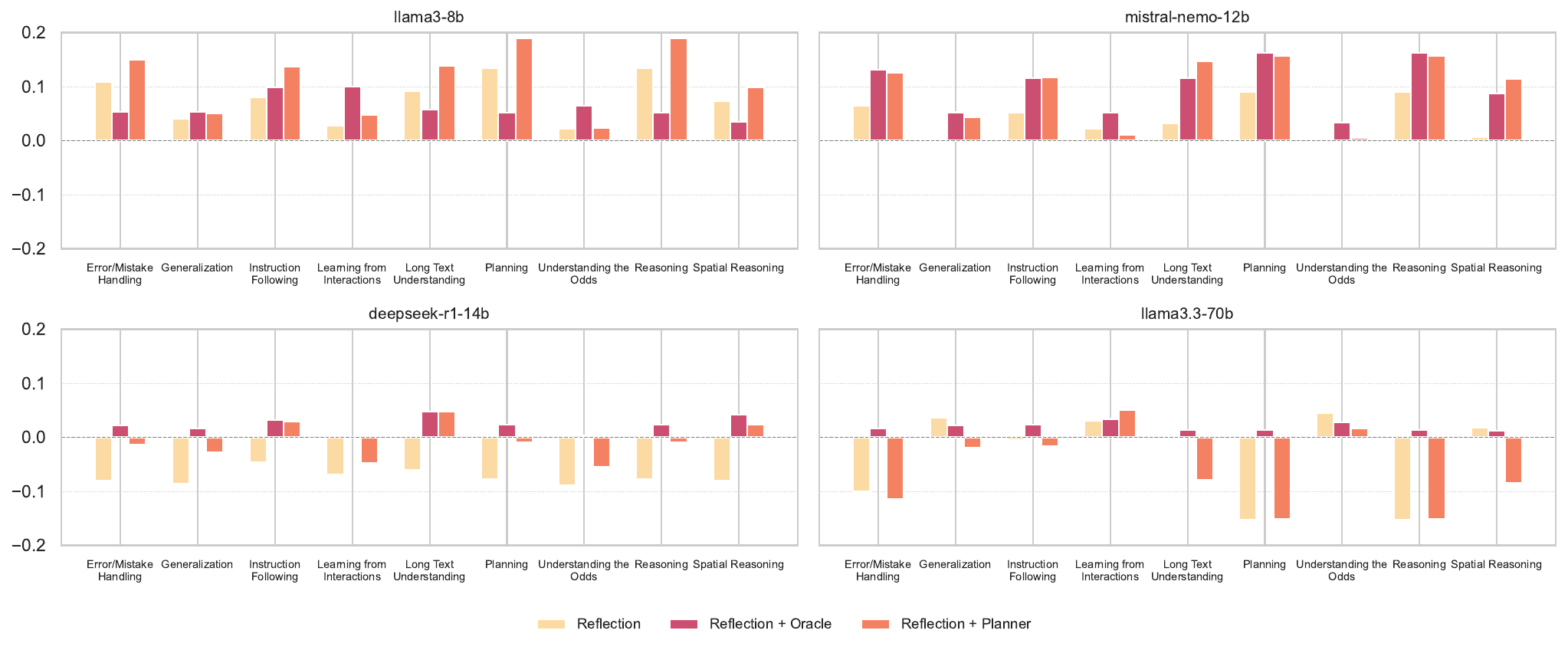}
\caption{
Normalized performance difference compared to baseline for various models and prompting strategies across nine challenges. \textsc{Llama3-8B} and \textsc{Mistral-Nemo-12B} generally benefit, especially from Reflection + Planner, with consistent improvements. In contrast, \textsc{DeepSeek-R1-14B} shows minimal gains, and Reflection can even degrade its performance. While some tasks improve slightly for \textsc{Llama3.3-70B}, many tasks show regressions (e.g., Planning score reduction with Reflection + Planner). All models consistently improve on Instruction Following. However, the overall impact of advanced prompting varies with model capacity and task, requiring tailored strategies for each model and challenge type.
}
  \label{fig:strategy-challenges}
\end{figure*}

\subsubsection{Strategies versus Challenge Type}
Figure~\ref{fig:strategy-challenges} shows the difference each prompting strategy achieves relative to the baseline on the nine SmartPlay skills.  For every game \(g\) we min--max scale the raw mean score \(r_{m,s,g}\) to \(x_{m,s,g}\in[0,1]\) (\(m=\)~model, \(s=\)~strategy).  The score difference is calculated as \(\delta_{m,s,g}=x_{m,s,g}-x_{m,\text{Base},g}\). SmartPlay assigns weights \(w_{g,d}\in\{0.33,0.67,1\}\) that indicate how strongly game \(g\) tests challenge dimension \(d\).  
Weighted aggregation then gives 
\begin{equation}
\Delta_{m,s,d} = 
\frac{\sum_{g} w_{g,d}\,\delta_{m,s,g}}{\sum_{g} w_{g,d}},
\label{eq:delta}
\end{equation}
where \(d\) indexes the challenge dimensions; a positive (negative) \(\Delta_{m,s,d}\) means strategy \(s\) increases (decreases) model \(m\)’s performance on dimension \(d\) relative to the baseline.

\textsc{Llama3-8B} benefits most from the Reflection + Planner strategy, achieving large gains (+0.19) on \textit{Planning} and \textit{Reasoning} challenges. The other two strategies also result in score increases compared to the baseline.  Similarly, \textsc{Mistral-Nemo-12B} shows improvements across most challenges, with the addition of an Oracle or Planner yielding greater gains than reflection alone. \textsc{DeepSeek-R1-14B} shows minimal improvements with prompting strategies: \textit{Long Text Understanding} shows a modest improvement of 0.05, and Reflection alone can even drag its performance down (for instance, –0.08 on \textit{Error Handling}). For the largest model, \textsc{Llama3.3-70B}, the structured prompts produce mixed results: they offer small gains in some challenges, such as improving \textit{Learning from Interactions} by +0.05 and Understanding the Odds by +0.04, but lead to adverse effects in others. For instance, the Reflection + Planner prompt reduces the \textit{Planning} score by approximately 0.15 relative to the baseline. All models consistently improve on Instruction Following. However, the impact of prompting strategies varies significantly across different models. While smaller models show improvements across all nine challenges, larger models show a regression on several challenges. This suggests that advanced prompting strategies can interfere with a high-capacity model's inherent abilities, highlighting the importance of tailoring the prompting strategy to the specific model and challenge. Next, we provide an analysis of the results per environment.

\subsubsection{TwoArmedBandit}
The optimal strategy in the two-armed bandit task relies on a basic count of past rewards from both machines. We observe that incorporating more complex prompting led to performance degradation compared to the baseline. For instance, \textsc{Llama3-8B}'s score drops from $40.35\,[37.45-41.65]$  to $34.00\,[30.00-35.00]$ with Reflection + Planner. Similarly, \textsc{DeepSeek-R1-14B} score drops from $41\,[40.55, 41.40]$ to $32.05\,[29.00,33.00]$ with Reflection + Planner. \textsc{LLAMA3.3-70B} is the only exception, which achieves a higher maximum score of $48.00$ when Reflection + Planner is enabled compared to the maximum Baseline score $41.90$.
Performance relies on rapid identification of the higher-paying bandit. The Base agent, exemplified by \textsc{Llama3-8b}, quickly exploits positive rewards, in contrast to other strategies that have lower performance by prolonged, oscillating exploration despite negative feedback (Example 1). 
Further inspection shows that Reflection, Oracle, and Planner discourage completely ignoring any single arm in favour of sustained exploration. Reflection points out to \textit{"avoid pulling the same machine too frequently without exploring the other option"}, while 
the Planner recommends alternating sequences like \textit{“Pull slot machine 2 → Pull slot machine 1 → Pull slot machine 2”}. This analysis reveals two complementary failure modes. First, Reflection, Oracle and Planner increase the agent's prompt length extensively, diluting important information and decreasing performance in smaller models. This is consistent with the idea that extraneous context reduces signal-to-noise ratio and adds \textit{distraction }\cite{tworkowski2023focused, liu2023lost}. Second, advanced reasoning methods encourage the agent to keep testing both arms even when one arm is clearly superior. We often observe the model \textit{overthinks} \cite{chen2024dont,stechly2024chain,fan2025missing} and reasons itself away from exploiting a profitable streak, leading to slower convergence.







\begin{figure}[!t] 
\begin{mainbox}{{\textsc{Llama3-8b} with Reflection and Oracle keeps exploring both bandits despite negative reward from slot machine 1.}}
\label{mybox:TwoArmedBandit_reflection_oracle}
Player Observation Step 1:
A new round begins.

Player Observation Step 2:
You pulled slot machine 1, you received reward -1.

Player Observation Step 3:
You pulled slot machine 2, you received reward 1.

Player Observation Step 4:
You pulled slot machine 1, you received reward -1.

Player Observation Step 5:
You pulled slot machine 2, you received reward 1.

Player Observation Step 6:
You pulled slot machine 1, you received reward -1.

\end{mainbox}
\vspace{\baselineskip} 
\end{figure}

\subsubsection{Rock Paper Scissors}
The highest gain appears for \textsc{Llama3.3-70B}: improving from $22.20\,[21.55,25.05]$ to $30.00\,[27.00-37.00]$ with Reflection + Planner.
\textsc{Llama3-8B} improves from $16.50\,[16.40-19.05]$ to  $26.00\,[12.00-36.00]$ with Reflection + Oracle. 
\textsc{Mistral-Nemo-12B} and \textsc{DeepSeek-R1-14B} do not see much improvement. Smaller models, especially \textsc{Llama3-8B} and \textsc{mistral-nemo-12b}, show occasional high-scoring runs but inconsistent overall play. 
Inspections show that the larger model shows better planning skills and adaptation to opponent patterns. \textsc{Llama3.3-70B} with Reflection + Planner reasoned that \textit{“choosing Paper has led to wins against Rock, which is a common choice”}, while \textit{“being prepared to adapt with Scissor if the opponent chooses Paper, and then switch to Rock if the pattern suggests an upcoming Scissor.”}. This ability to vary moves and execute planned sequences highlights a strong connection between reasoning and action. 
Smaller models often fell into repetition. 
In one episode, \textsc{Llama3-8B} agent opens with \textit{Paper} and then continues to play \textit{Paper} in most rounds, even while the opponent frequently plays \textit{Scissors}. The agent incurs a string of losses but does not switch strategy. These cases show the smaller models failing to adjust even when a pattern is not working, leading to exploitable behaviour. 



\subsubsection{Tower of Hanoi}

\textsc{Llama3.3-70B} achieves the best performance in Tower of Hanoi, with a score of 2.00 [2.00-2.00], matched by Reflection + Oracle. However, its score drops to 1.00 [1.00-2.00] with Reflection + Planner, and to 0.70 [1.00-1.80] with Reflection alone. While for smaller models, advanced prompting can significantly improve performance, this improvement is brittle and can lead to performance drops. \textsc{DeepSeek-R1-14B} has a baseline score of 1.00 [0.60-1.20]. With Reflection + Planner, its score fluctuates, reaching as high as 2.00 but also dropping to 1.00. Similarly, for \textsc{Llama3-8B}, the baseline score of 0.20 [0.20-0.30] has lower variability, yet Reflection and Planner can achieve a maximum score of 2.00, though it may also drop to a minimum of 0.00. 
\textsc{Mistral-Nemo-12B} moves from $0.00\,[0.00- 0.00]$ to $1.00\,[0.00-1.00]$ with both Reflection + Oracle and Reflection + Planner.
Compared with other 
tasks, agents using Planner struggle more to interpret the disk-placement state. 
Hanoi failures frequently resulted from invalid moves, such as placing a larger disk onto a smaller one (Example 2). 
The agent's reflection identified this recurring problem: \textit{"Based on the trajectory observed, there's a common pitfall that has been repeatedly encountered: attempting to place a bigger disk on top of a smaller one."}. While the models can recite the optimal seven-move sequence for the Tower-of-Hanoi when prompted, the agents struggled, averaging 30 moves without completion, repeatedly shuffling disks back and forth or performing invalid moves. This illustrates a broader pattern: LLMs have seen Tower-of-Hanoi in training and can complete prompts about its optimal solution, yet they do not internalise the rules well enough to act on it. Another challenge is the sparse reward function: 0 for legal moves, -1 for illegal moves, and +100 upon completion, which gives the agent little positive feedback before the puzzle is solved.

\begin{figure}[!t] 
  \centering
  \begin{minipage}{\linewidth}    
\begin{mainbox}{\textsc{Mistral-Nemo-12b} made an invalid move in Tower of Hanoi by attempting to place a larger disk (size 2 from rod B) onto a smaller one (size 0 on rod C).}
\label{mybox:invalid_action_hanoi}

\textbf{State (Before Action):}

- A: |bottom, [], top|

- B: |bottom, [2], top|

- C: |bottom, [1, 0], top|

\textbf{Attempted Action:} Move the top disk from rod B to rod C.

\textbf{Reward:} -1 (Indicates an invalid move)

\textbf{Next State (Unchanged due to invalid action):}

- A: |bottom, [], top|

- B: |bottom, [2], top|

- C: |bottom, [1, 0], top|
\end{mainbox}
  \end{minipage}
    \vspace{\baselineskip} 
\end{figure}







\begin{table*}[!tb]
    \centering
    \caption{\textsc{Mistral-Nemo-12b} performance over five runs on Hanoi across four conditions and a random-action baseline. Values show the percentage of episodes where agents achieved the goal (G), average disks placed on target peg (D), and percentage of invalid moves (I). Darker colours mean better performance across the column. In the 3-disk task, showing valid actions proved more beneficial than reward shaping for disk placement and reducing invalid moves; combining them offered a minor boost to goal achievement. Surprisingly, the random policy outperforms all methods in the no adjustments setting across goal, disk placement, and invalid‑move metrics. For 2 disks, all methods achieve substantially higher success than in the 3-disk scenario, with valid‑action hints most effective at achieving the goal and reducing invalid moves. Combining shaping and showing valid actions yields no consistent advantage, indicating the main challenge is not the reward signal but identifying valid moves.}
    \label{table:hanoi_stats}
    \scriptsize
    \begin{tabular}{l|ccc|ccc|ccc|ccc}
        \toprule
        \multirow{2}{*}{Method} & \multicolumn{3}{c}{No adjustments} & \multicolumn{3}{c}{Reward Shaping} & \multicolumn{3}{c}{Show Valid Actions} & \multicolumn{3}{c}{\shortstack{Show Valid Actions\\+ Reward Shaping}} \\
        \cmidrule{2-13}
        & G & D & I & G & D & I & G & D & I & G & D & I \\
        \midrule
        \multicolumn{13}{c}{\textbf{3 Disks}} \\
        \midrule
        Base & \cellcolor[HTML]{FCFDBF}0.0\% & \cellcolor[HTML]{FCFDBF}0.2 & \cellcolor[HTML]{FCFDBF}79.4\% & \cellcolor[HTML]{FCFDBF}0.0\% & \cellcolor[HTML]{FCC08D}0.4 & \cellcolor[HTML]{FCF8BB}79.0\% & \cellcolor[HTML]{FCAF80}2.0\% & \cellcolor[HTML]{B7371B}1.1 & \cellcolor[HTML]{B7371B}56.3\% & \cellcolor[HTML]{FCFDBF}0.0\% & \cellcolor[HTML]{D55A39}0.9 & \cellcolor[HTML]{BE4023}57.6\% \\
        Reflection & \cellcolor[HTML]{FCFDBF}0.0\% & \cellcolor[HTML]{FCA275}0.6 & \cellcolor[HTML]{FCC793}74.1\% & \cellcolor[HTML]{FCAF80}2.0\% & \cellcolor[HTML]{D55A39}0.9 & \cellcolor[HTML]{FCA376}70.5\% & \cellcolor[HTML]{FCFDBF}0.0\% & \cellcolor[HTML]{E46C48}0.8 & \cellcolor[HTML]{D45A39}61.3\% & \cellcolor[HTML]{FCAF80}2.0\% & \cellcolor[HTML]{E16845}0.8 & \cellcolor[HTML]{D95F3D}62.0\% \\
        Reflection + Oracle & \cellcolor[HTML]{FCFDBF}0.0\% & \cellcolor[HTML]{F6815A}0.7 & \cellcolor[HTML]{FCC793}74.1\% & \cellcolor[HTML]{FCFDBF}0.0\% & \cellcolor[HTML]{EA734E}0.8 & \cellcolor[HTML]{FCA376}70.5\% & \cellcolor[HTML]{E56D49}4.0\% & \cellcolor[HTML]{CC4F30}1.0 & \cellcolor[HTML]{DD6542}62.8\% & \cellcolor[HTML]{FCAF80}2.0\% & \cellcolor[HTML]{DE6542}0.9 & \cellcolor[HTML]{D85F3D}62.0\% \\
        Reflection + Planner & \cellcolor[HTML]{FCFDBF}0.0\% & \cellcolor[HTML]{FCCF9A}0.4 & \cellcolor[HTML]{FCDDA5}76.3\% & \cellcolor[HTML]{FCFDBF}0.0\% & \cellcolor[HTML]{FCCF9A}0.4 & \cellcolor[HTML]{FCE2A9}76.8\% & \cellcolor[HTML]{FCFDBF}0.0\% & \cellcolor[HTML]{D55A39}0.9 & \cellcolor[HTML]{FCE8AE}77.4\% & \cellcolor[HTML]{FCFDBF}0.0\% & \cellcolor[HTML]{DE6542}0.9 & \cellcolor[HTML]{FCE7AD}77.2\% \\
        Random & \cellcolor[HTML]{FCAF80}2.0\% & \cellcolor[HTML]{D85E3C}0.9 & \cellcolor[HTML]{FC9B70}69.7\% & \cellcolor[HTML]{FCFDBF}0.0\% & \cellcolor[HTML]{ED7751}0.8 & \cellcolor[HTML]{FC9F73}70.1\% & \cellcolor[HTML]{FCFDBF}0.0\% & \cellcolor[HTML]{D85E3C}0.9 & \cellcolor[HTML]{FC9369}68.9\% & \cellcolor[HTML]{B7371B}6.0\% & \cellcolor[HTML]{D55A39}0.9 & \cellcolor[HTML]{FC966B}69.2\% \\
        \midrule
        \multicolumn{13}{c}{\textbf{2 Disks}} \\
        \midrule
        Base & \cellcolor[HTML]{FCFDBF}2.0\% & \cellcolor[HTML]{FCFDBF}0.5 & \cellcolor[HTML]{FCD9A2}72.0\% & \cellcolor[HTML]{FCCD98}16.0\% & \cellcolor[HTML]{FCBE8C}0.8 & \cellcolor[HTML]{FCD6A0}71.6\% & \cellcolor[HTML]{EB7550}44.0\% & \cellcolor[HTML]{E06845}1.2 & \cellcolor[HTML]{DB623F}50.1\% & \cellcolor[HTML]{EB7550}44.0\% & \cellcolor[HTML]{DB6240}1.3 & \cellcolor[HTML]{D75E3C}49.1\% \\
        Reflection & \cellcolor[HTML]{FCBF8D}20.0\% & \cellcolor[HTML]{FCBA89}0.8 & \cellcolor[HTML]{FCCC97}69.9\% & \cellcolor[HTML]{FC9D71}30.0\% & \cellcolor[HTML]{FC8D64}1.0 & \cellcolor[HTML]{FCC894}69.4\% & \cellcolor[HTML]{BF4023}66.0\% & \cellcolor[HTML]{C34527}1.5 & \cellcolor[HTML]{C6482A}44.3\% & \cellcolor[HTML]{B7371B}70.0\% & \cellcolor[HTML]{B7371B}1.6 & \cellcolor[HTML]{B7371B}40.2\% \\
        Reflection + Oracle & \cellcolor[HTML]{EF7A54}42.0\% & \cellcolor[HTML]{F9865E}1.0 & \cellcolor[HTML]{FCBA88}66.9\% & \cellcolor[HTML]{EF7A54}42.0\% & \cellcolor[HTML]{F9865E}1.0 & \cellcolor[HTML]{FCB988}66.9\% & \cellcolor[HTML]{BB3B1F}68.0\% & \cellcolor[HTML]{BE3F22}1.5 & \cellcolor[HTML]{BF4124}42.6\% & \cellcolor[HTML]{DB6240}52.0\% & \cellcolor[HTML]{E36B48}1.2 & \cellcolor[HTML]{E26B47}52.1\% \\
        Planner & \cellcolor[HTML]{FCAB7C}26.0\% & \cellcolor[HTML]{F9865E}1.0 & \cellcolor[HTML]{FCE9AF}74.7\% & \cellcolor[HTML]{FCE1A8}10.0\% & \cellcolor[HTML]{FCA175}0.9 & \cellcolor[HTML]{FCFDBF}77.8\% & \cellcolor[HTML]{DB6240}52.0\% & \cellcolor[HTML]{C5482A}1.5 & \cellcolor[HTML]{FCA578}63.7\% & \cellcolor[HTML]{FCAE7F}25.0\% & \cellcolor[HTML]{EB7550}1.1 & \cellcolor[HTML]{FCDCA4}72.5\% \\
        Random & \cellcolor[HTML]{FC966C}32.0\% & \cellcolor[HTML]{FCBA89}0.8 & \cellcolor[HTML]{FCC390}68.5\% & \cellcolor[HTML]{FC8F66}34.0\% & \cellcolor[HTML]{FC8D64}1.0 & \cellcolor[HTML]{FCC793}69.1\% & \cellcolor[HTML]{D35837}56.0\% & \cellcolor[HTML]{CF5434}1.4 & \cellcolor[HTML]{FCA276}63.2\% & \cellcolor[HTML]{EF7A54}42.0\% & \cellcolor[HTML]{E8714D}1.2 & \cellcolor[HTML]{FCB786}66.5\% \\
        \bottomrule
    \end{tabular}
\end{table*}

\begin{table*}[!tb]
    \centering
    \caption{\textsc{Mistral-Nemo-12b} 
 performance over five runs on Messenger across four conditions and a random-action baseline. Values show the percentage of episodes where agents successfully picked up messages (P), achieved goals (G), or collided with enemies (C). Darker colours mean better performance across the column. Removing synonyms only slightly improves the pickup rates and the goal achievement. Reward shaping has a strong positive effect on the pickup rate but not on goal achievement. Combining both does not yield significant better results than reward shaping alone. }
    \label{table:messenger_stats}
    \scriptsize
    \begin{tabular}{l|ccc|ccc|ccc|ccc}
        \toprule
        \multirow{2}{*}{Method} & \multicolumn{3}{c}{No adjustments} & \multicolumn{3}{c}{No Synonyms} & \multicolumn{3}{c}{Reward Shaping} & \multicolumn{3}{c}{\shortstack{Reward Shaping\\+ No synonyms}} \\
        \cmidrule{2-13}
        & P & G & C & P & G & C & P & G & C & P & G & C \\
        \midrule
        Base & \cellcolor[HTML]{C94D2D}7.0\% & \cellcolor[HTML]{C94D2D}7.0\% & \cellcolor[HTML]{FCA376}32.0\% & \cellcolor[HTML]{D25736}8.0\% & \cellcolor[HTML]{D25736}8.0\% & \cellcolor[HTML]{FCFDBF}36.0\% & \cellcolor[HTML]{FC9168}21.5\% & \cellcolor[HTML]{FC9067}4.0\% & \cellcolor[HTML]{FC8C63}35.5\% & \cellcolor[HTML]{FCE9AE}15.0\% & \cellcolor[HTML]{FC9C70}5.0\% & \cellcolor[HTML]{FCFDBF}37.0\% \\
        Reflection & \cellcolor[HTML]{FCFDBF}2.0\% & \cellcolor[HTML]{FCFDBF}2.0\% & \cellcolor[HTML]{FCFDBF}37.0\% & \cellcolor[HTML]{FCBF8C}4.0\% & \cellcolor[HTML]{FCBF8C}4.0\% & \cellcolor[HTML]{FCE6AD}35.5\% & \cellcolor[HTML]{FCFABC}16.5\% & \cellcolor[HTML]{B7371B}6.5\% & \cellcolor[HTML]{BB3B1F}27.5\% & \cellcolor[HTML]{FCC18E}21.0\% & \cellcolor[HTML]{FCE9AF}3.0\% & \cellcolor[HTML]{EF7A54}27.0\% \\
        Reflection + Oracle & \cellcolor[HTML]{F6825B}5.1\% & \cellcolor[HTML]{F6825B}5.1\% & \cellcolor[HTML]{C84B2C}25.6\% & \cellcolor[HTML]{FCA175}4.9\% & \cellcolor[HTML]{FCA175}4.9\% & \cellcolor[HTML]{B7371B}30.8\% & \cellcolor[HTML]{FCFDBF}16.4\% & \cellcolor[HTML]{FCFDBF}1.8\% & \cellcolor[HTML]{E36B47}32.3\% & \cellcolor[HTML]{FCD19C}18.5\% & \cellcolor[HTML]{FCFDBF}2.5\% & \cellcolor[HTML]{E36C48}25.5\% \\
        Reflection + Planner & \cellcolor[HTML]{B7371B}7.8\% & \cellcolor[HTML]{B7371B}7.8\% & \cellcolor[HTML]{FCCF9A}34.4\% & \cellcolor[HTML]{B7371B}9.5\% & \cellcolor[HTML]{B7371B}9.5\% & \cellcolor[HTML]{E46D49}32.5\% & \cellcolor[HTML]{B7371B}27.5\% & \cellcolor[HTML]{D45A38}5.5\% & \cellcolor[HTML]{FCFDBF}43.5\% & \cellcolor[HTML]{B7371B}47.0\% & \cellcolor[HTML]{B7371B}8.5\% & \cellcolor[HTML]{F37F58}27.5\% \\
        Random & \cellcolor[HTML]{F9855E}5.0\% & \cellcolor[HTML]{F9855E}5.0\% & \cellcolor[HTML]{B7371B}24.0\% & \cellcolor[HTML]{FCFDBF}2.0\% & \cellcolor[HTML]{FCFDBF}2.0\% & \cellcolor[HTML]{F17C56}33.0\% & \cellcolor[HTML]{FCC692}19.0\% & \cellcolor[HTML]{FCC28F}3.0\% & \cellcolor[HTML]{B7371B}27.0\% & \cellcolor[HTML]{FCFDBF}12.0\% & \cellcolor[HTML]{FCC390}4.0\% & \cellcolor[HTML]{B7371B}20.0\% \\
        \bottomrule
    \end{tabular}
\end{table*}

\subsubsection{Messenger}
Smaller models gain from Reflection + Planner in Messenger. \textsc{Llama3-8B}'s score improves from 
$-0.15\,[-0.45- -0.15)$ to $0.00\,[0.00-0.00]$,
\textsc{Mistral-Nemo-12B} improves from  $-0.20\,[-0.20- -0.15)$ to $1.00\,[-1.00-1.00]$, and \textsc{DeepSeek-R1-14B} improves from $0.40[0.25-0.55]$ to $1.00[0.40-1.00)$. The same Reflection + Planner collapses \textsc{Llama3.3-70B} from $0.10\,[-0.10-0.10]$ to $-1.00\,[-1.00-1.00]$. This decline comes from the planner's strategy, leading to overly cautious enemy avoidance, resulting in detours and failure to complete the task within ten steps.
Object misidentification and poor spatial awareness frequently led to agent failures in Messenger. 
For example, \textsc{Llama3-8b}'s heuristics fail to adapt to the changing positions of enemy, message, and goal, often referring to objects that are not present and thereby confusing the agent. \textsc{Mistral-Nemo-12b} once incorrectly identified an \textit{airplane} as the goal in one episode. 
Although the agent later attempted self-correction, stating, \textit{"the airplane, which is the enemy, is one step away and not moving. To avoid losing immediately, we should move towards the message (dog) located three steps southwest"}, its subsequent eastward move resulted in an enemy encounter. 
\textsc{DeepSeek-R1-14B} and \textsc{LLAMA3.3-70B} demonstrate better synonym comprehension, and the latter is also better in creating  generalizable heuristics: \textit{"Identify and move towards objects that match the description of a "message" (e.g., "classified report", "restricted document"), as obtaining the message is crucial to winning the game," and "Exercise caution around objects described with synonyms of "danger" or "enemy."}. Similar to Hanoi, the reward function is not very informative. Positive outcomes only occur upon picking up the message or reaching the goal, events which happen infrequently. Most actions offer no reward, and colliding with enemies ends the episode early.

\subsection{Additional Analysis}

We modify the setups for \textsc{Hanoi3Disks} and \textsc{MessengerL1} to investigate failure modes and potential improvements. The experiments are conducted with \textsc{Mistral-Nemo-12b} to balance speed and performance.

\subsubsection{Hanoi}

We implement the following modifications: simplify the puzzle to two disks, add valid actions hints, and reward shaping (-2 for invalid, +1 for valid moves, +100 for goal). 
The results over five runs are summarized in Table \ref{table:hanoi_stats}.

Under the baseline 3-disk setting, the agent never completes the puzzle and makes frequent illegal moves, typically ending episodes with one disk on the target. With reward shaping, the agent places slightly more disks correctly and commits fewer illegal moves. Showing valid actions reduces illegal moves but does not eliminate them, likely because the hints get lost in the lengthy prompt. The Planner often becomes confused by the puzzle’s intermediate states and stops choosing valid moves altogether. Combining reward shaping with valid-action hints yields no benefit beyond showing valid actions alone. \textsc{Mistral-Nemo-12B} underperforms even a uniform-random policy on success rate, correct placements, and move legality, revealing a fundamental inability to internalise the 3-disk Tower-of-Hanoi rules. Under the two-disk scenario, all methods achieve higher goal completion rates than in the three-disk setting. Without additional modifications, \textit{Reflection + Oracle} attains 42\% goal success. Reward shaping boosts performance for most methods, except the Planner, while showing valid actions further reduces illegal moves and raises completion rates. However, combining reward shaping with showing valid actions does not consistently produce better outcomes, as in the 3-disk setting, indicating that LLMs can recall short move chains (2-disk) and that the actual bottleneck is the
LLM’s lack of inherent understanding of the game’s core
constraint rather than a shortage of feedback.

\subsubsection{Messenger}
We apply reward shaping for Messenger: incremental rewards for moving closer to the message or goal, and larger bonuses for message pickup (increased from 1.0 to 10.0) and final delivery (from 1.0 to 50.0). We introduce small distance-based rewards (+0.50 per step closer to the message or goal while carrying it), producing a denser reward signal to guide the agent. We also remove object synonyms to isolate the impact of language complexity. The results over five runs are summarized in Table \ref{table:messenger_stats}. Synonyms removal yields marginal gains in goal completion and pickup rate for all methods, except for Reflection + Planner, where the pickup rate and goal achievement drop from 5.1\% to 4.9\%. Reward shaping consistently boosts pickup rates across all methods, but does not improve goal attainment. When reward shaping and synonym removal are combined, the pickup rate is higher, and the collision rate is lower than in the condition with no adjustments (except for the baseline). Reflection + Planner with reward shaping and synonym removal gives the best overall result (P = 47\%, G = 8.5\%, collisions = 27\%). Despite reward shaping and synonym removal, high collision rates persist, indicating a fundamental spatial awareness and navigation limitation.
Messenger exposes a gap between language comprehension and embodied competence: language-only pre-training equips LLMs with object labels and simple heuristics but not the grounded, stateful planning needed for dynamic spatial tasks. Bridging this gap will likely require tighter coupling between language and perception rather than prompt engineering.

\section{Conclusion and Discussion}
\label{conclusion}

Our evaluation of open-source LLMs on the SmartPlay benchmark reveals both the potential and limitations of advanced prompting strategies, including self-reflection, heuristic mutation, and planning. We find that excessive reasoning harms the performance of smaller models on simple tasks, as it forces the model to filter through more content and increase the signal-to-noise ratio \cite{tworkowski2023focused}. This reasoning not only \textit{distracts }\cite{tworkowski2023focused} but also causes the models to \textit{overthink} \cite{chen2024dont}, leading the model to overcomplicate the process and disregard simpler, more effective solutions. In addition, we find that larger models perform better, but that strategic prompting can close this gap, and that having a dense, task-aligned reward signal can improve an agent’s decision-making, offering a simpler alternative compared to the significant effort needed to find optimal prompts. While smaller models particularly benefit from advanced prompting in complex tasks, our results demonstrate significant variability: the same prompt can produce substantial gains or, conversely, lead to worse performance than the baseline; this holds for all model sizes. This highlights the need for more robust solutions. Moreover, the common practice in reasoning studies of reporting aggregate performance metrics, such as accuracy  \cite{wei2022chain} or F1 scores \cite{yang2018hotpotqa}
can be misleading and obscure the sensitivity of results to variations caused by instability, affecting the generalization and reproducibility of findings. 
Finally, while LLMs might exhibit proficiency on in-distribution data, we find little evidence for emergent reasoning. We find that LLMs expose a knowing-doing gap\cite{paglieri2024balrog}: models can recite optimal solution learning during training (e.g. the seven-move Tower of Hanoi) but fail to execute them in practice. We also find a language-embodiment gap: textual understanding does not translate into grounded, stateful planning. Our findings support the growing evidence necessitating a critical examination of prompting methods claiming emergent LLM abilities and highlight the need to re-evaluate current benchmarks like question-answer pairs or math word problems, which inadequately capture the complexity of reasoning and fail to reveal inherent deficits. 
Future work could benefit from combining in-context learning with external memory to improve recall, symbolic abstractions to ensure verifiable reasoning, denser task‑aligned feedback, and multimodal perception to ground the agent's understanding in the physical world.




\bibliographystyle{IEEEtran}
\bibliography{main.bib}

\end{document}